\documentclass[runningheads]{llncs}
\usepackage{graphicx}
\usepackage{amsmath,amssymb} % define this before the line numbering.
\usepackage{color}
\usepackage[width=122mm,left=12mm,paperwidth=146mm,height=193mm,top=12mm,paperheight=217mm]{geometry}
\DeclareMathOperator{\E}{\mathbb{E}}
\usepackage{multirow}
\usepackage[pagebackref=true,breaklinks=true,letterpaper=true,colorlinks,bookmarks=false]{hyperref}

\usepackage{wasysym}

\newcommand{\figref}[1]{Figure~\ref{#1}}
\newcommand{\tabref}[1]{Table~\ref{#1}}

\newcommand{\secref}[1]{Section~\ref{#1}}
\newcommand{\comment}[1]{{}}

\newcommand{\eg}{\textit{e.g.}}
\newcommand{\etal}{\textit{et al.}}

\newcommand{\bigcell}[2]{\begin{tabular}{@{}#1@{}}#2\end{tabular}}

\begin{document}
% \renewcommand\thelinenumber{\color[rgb]{0.2,0.5,0.8}\normalfont\sffamily\scriptsize\arabic{linenumber}\color[rgb]{0,0,0}}
% \renewcommand\makeLineNumber {\hss\thelinenumber\ \hspace{6mm} \rlap{\hskip\textwidth\ \hspace{6.5mm}\thelinenumber}}
% \linenumbers
\pagestyle{headings}
\mainmatter

\title{Fine-scale Surface Normal Estimation using a Single NIR Image} % Replace with your title

\titlerunning{Fine-scale Surface Normal Estimation using a Single NIR Image}

\authorrunning{Yoon et. al.}

\author{Youngjin Yoon$^1$$\;\;\;\;\;\;\;\;\;\;\;\;\;\;\;$Gyeongmin Choe$^1$$\;\;\;\;\;\;\;\;\;\;\;\;\;\;\;$Namil Kim$^1$\\$\;\;\;\;${\small yjyoon@rcv.kaist.ac.kr}$\;\;\;\;\;\;\;\;${\small gmchoe@rcv.kaist.ac.kr}$\;\;\;\;\;\;\;${\small nikim@rcv.kaist.ac.kr}$\;\;$\\Joon-Young Lee$^2$$\;\;\;\;\;\;\;\;\;\;\;\;\;\;\;\;\;$In So Kweon$^1$\\$\;\;\;\;\;\;\;${\small jolee@adobe.com}$\;\;\;\;\;\;\;\;\;\;\;\;\;\;\;${\small iskweon@kaist.ac.kr}$\;\;\;$}\institute{$^1$KAIST\;\;\;$^2$Adobe Research.}

\maketitle

\begin{abstract}
We present surface normal estimation using a single near infrared (NIR) image.
We are focusing on fine-scale surface geometry captured with an uncalibrated light source.
To tackle this ill-posed problem, we adopt a generative adversarial network which is effective in recovering a sharp output, which is also essential for fine-scale surface normal estimation.
We incorporate angular error and integrability constraint into the objective function of the network to make estimated normals physically meaningful.
We train and validate our network on a recent NIR dataset~\cite{Choe16cvpr}, and also evaluate the generality of our trained model by using new external datasets which are captured with a different camera under different environment.
\keywords{Shape from shading, near infrared image, generative adversarial network}
\end{abstract}

\section{Introduction}

Estimating surface geometry is a fundamental problem to understand the properties of an object and reconstruct 3D information of it.
There are two different approaches; geometric methods such as structure-from-motion and multi-view stereo, and photometric methods such as photometric stereo and shape-from-shading. The geometric methods are usually useful for metric reconstruction while the photometric methods are effective for estimating accurate per-pixel surface geometry.

Recently, with the massive use of commercial depth sensors, \eg, Kinect and RealSense, many works have been proposed to enhance the depth quality of the sensors by fusing the photometric cues of a color image~\cite{han2013high,cvpr13_kinectrefinement} or a near infrared (NIR) image~\cite{choe2014exploiting,haque2014high}. Although the methods have proven the effectiveness of photometric shape estimation and have provided promising results, they rely highly on the sensors and usually require heavy computational time.

On the other hand, deep convolutional neural networks (CNN) have been broadly used for various computer vision tasks such as image classification~\cite{he2015deep,krizhevsky2012imagenet}, object detection~\cite{yoo2015attentionnet,girshick2014rich}, segmentation~\cite{long2015fully,hong2015decoupled}, and depth estimation~\cite{li2015depth,eigen2015predicting}.
With its rich learning capability, deep CNN has shown the state-of-the-art performances in many areas and also make the algorithms more practical with fast evaluation time.
Lately, several works have also tried to solve depth or surface normal estimation using CNN~\cite{li2015depth,eigen2015predicting}. However, they mostly focused on scene-level estimation~\cite{eigen2015predicting} or context-aware methods~\cite{wang2015designing} which generate rough surface normals, therefore they cannot generate the fine-scale surface details of a target object.

\begin{figure}[t]
	\centering
	\begin{tabular}{c@{\hspace{1mm}}c@{\hspace{1mm}}}
		\includegraphics[width=0.99\linewidth]{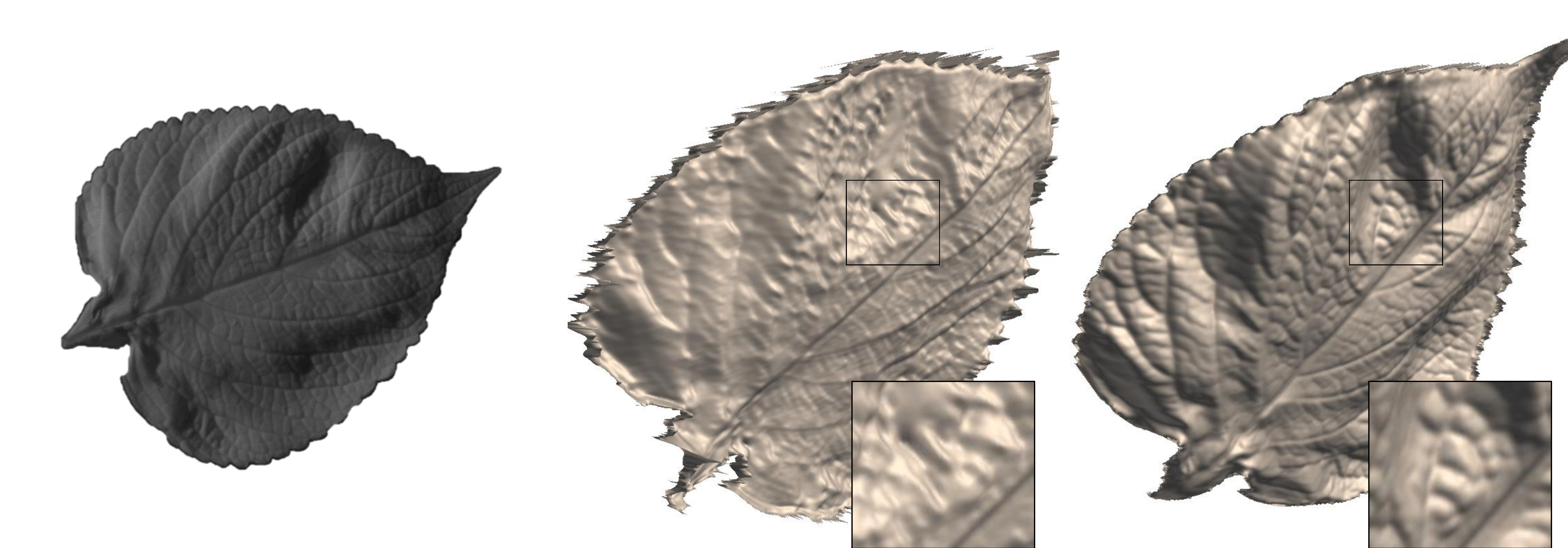}
	\end{tabular}
	\caption{Comparison of reconstruction results, left: Input NIR image, middle: Our reconstruction from a single NIR image, right: ground-truth reconstruction using NIR images captured under 12 different lighting directions.}
	\label{fig:surface}
\end{figure}

In this paper, we are mainly focusing on building a practical system for estimating fine-scale surface geometry, therefore we tackle the shape-from-shading problem with an uncalibrated light source. 
We solve the problem by training a deep CNN architecture on a recent NIR dataset~\cite{Choe16cvpr}. 
This dataset consists of 101 objects, captured by a NIR camera with 9 different viewing directions and 12 lighting directions. It allows us to train variety of textures such as fabrics, leaves, and papers. 
We propose to train the mapping between NIR intensity distributions and normal map with a generative adversarial network (GAN). 
We design the objective function of the GAN model to consider photometric characteristics of surface geometry by incorporating angular error and integrability constraint. Since we train various object images captured from different lighting directions, our method estimates fine-scale surface normals without the calibration of lighting direction. 
We verify that deep CNN is effective in handling the ill-posed uncalibrated shape-from-shading problem without complex heuristic assumptions.
Also, we evaluate the generality of our trained model by testing our own datasets which are captured using different configuration compared to the training dataset.
One example result of our method is shown in \figref{fig:surface}.

The major contributions of our work are as follows:
\begin{enumerate}
	\item First work analyzing the relationship between a NIR image and surface normal using a deep learning framework.
	\item Fine-scale surface normal estimation using a single NIR image where the light direction needs not be calibrated.
	\item Suitable design of an object function to reconstruct the fine details of a target object surface.
\end{enumerate}

%We demonstrate the practicality of our method using various real-world examples.   

\section{Related Work}

\subsubsection{Photometric Stereo and Shape from Shading}
Photometric stereo~\cite{woodham1980photometric} is one of the well-studied methods for estimating surface normals. 
By taking at least 3 images captured under different lighting directions, photometric stereo can uniquely determine the surface normals of an object. Also, the more usage of images make the output more accurate since it becomes an over-determined problem.

Shape from shading is a special case of photometric stereo, which predicts a shape from a single image. This is an ill-posed problem and needs to exploit many restrictions and constraints~\cite{zheng1991estimation,barron2012shape}. Begin with numerical SfS methods~\cite{ikeuchi1981numerical}, many works have shown results based on the Lambertian BRDF assumption. Tsai\etal~\cite{ping1994shape} use discrete approximation of surface normals. Lee and Kuo~\cite{lee1993shape} estimate shape by using a triangular element surface model. We refer readers to \cite{zhang1999shape} for better understanding regarding comparisons and evaluations of the classical SfS methods.

Shape from a NIR image has been recently studied in several literatures~\cite{choe2014exploiting,Choe16cvpr}. They analyze the discriminative characteristics of NIR images and experimentally show the albedo (surface reflectance) simplicity in the NIR wavelength of various materials. In \cite{choe2014exploiting,haque2014high}, they propose the shape refinement methods using the photometric cues in NIR images. They show the high-quality shape recovery results, however they need an additional depth camera to obtain the results. 

Although many conventional photometric approaches can work on NIR images and the albedo simplicity in the NIR image actually help robust estimation, estimating the surface normal from a single NIR image still have many limitations for practical uses, such as heavy computation time, heuristic assumptions, special system configuration, and the calibration of a light direction. To overcome those limitations, we studies the mapping from NIR intensity distributions to surface normal vectors via a deep CNN framework. We combine generative adversarial networks~\cite{goodfellow2014generative} with the specially designed objective function. Through the adversarial training process, our network naturally encodes the photometric cues of a scene and produces fine surface normals.

\subsubsection{Data-Driven Shape estimation}
There have been various studies on estimating the shape information from images via data-driven approaches.
Saxena~\etal~\cite{saxena2005learning} estimate depths using a discriminatively trained MRF model with multiple scales of monocular cues. Hoiem~\etal~\cite{hoiem2005automatic} reconstruct rough surface orientations of a scene by statistically modeling categories of coarse structures (\eg, ground, sky and vertical). 
Ladicky~\etal~\cite{ladicky2014pulling} incorporate semantic labels of a scene to predict better depth outputs.

One of the emerging directions for shape estimation is using deep CNN. In~\cite{fouhey2013data}, Fouhey~\etal try to discover the right primitives in a scene. 
In~\cite{wang2015designing}, Wang~\etal explore the effectiveness of CNNs for the tasks of surface normal estimation. Although this work infers the surface normals from a single color image, it outputs scene-level rough geometries and is not suitable for object-level detailed surface reconstruction. To estimate the object shape and the material property, Rematas~\etal~\cite{rematas2015deep} use the two different CNN architectures which predict surface normals directly and indirectly. The direct architecture estimates a reflectance map from an input image while the indirect architecture estimates a surface orientation map as an intermediate step towards reflectance map estimation.
In~\cite{liu2015deep}, Liu~\etal estimate depths from a single image using a deep CNN framework by jointly learning the unary and pairwise potentials of the CRF loss.  
In~\cite{eigen2014depth}, Eigen~\etal use a multi-scale approach which uses coarse and fine networks to estimate a better depth map.

Compared to the existing works, we focus on estimating fine-scale surface normals suing a deep CNN framework, therefore we bear in mind to design a network to produce photometrically meaningful outputs.

\section{Method}

\subsection{Generative Adversarial Network}
\label{sec:gan}
Generative adversarial network (GAN)~\cite{goodfellow2014generative} is a framework for training generative models which consists of two different models; a generative network $G$ for modeling the data distribution and a discriminative network $D$ for estimating the state of a network input. Therefore, the generative network learns to generate more realistic images making the latter to misjudge, while the discriminative network learns to correctly classify its input into a real image and a generated image. The two networks are simultaneously trained through a minimax optimization. 

Given an input image of the discriminative network, an initial discriminative parameter $\theta_D$ is stochastically updated to correctly predict whether the input comes from a training image $I$ or a generated image $F$. After that, with keeping discriminative parameter $\theta_D$ fixed, a generative parameter $\theta_G$ is trained to produce the better quality of images, which could be misclassified by the discriminative network as real images. These procedures are repeated until they converge. This minimax objective is denoted as:
\begin{equation}
	\min_{\Theta_G}\max_{\Theta_D}\\
	\displaystyle \E_{F\sim D_{desire}}[logD( I )] + \\
	\displaystyle \E_{Z\sim D_{input}}[log(1 - D(F) )]
\end{equation}
where $D_{desire}$ is the distribution of images that we desired to estimate and $D_{input}$ is that of the input domain. This objective function encourages $D$ to be assigned to the correct label to both real and generated images and make $G$ generate a realistic output $F$ from an input $Z$. In our method, both the generative and the discriminative model are based on convolutional networks. The former takes a single NIR image as an input and results in a three-dimensional normal image as an output. The latter classifies an input by using the binary cross-entropy to make the probability high when an input comes from the training data.

\begin{figure}[t]
\centering
\includegraphics[width=1\linewidth]{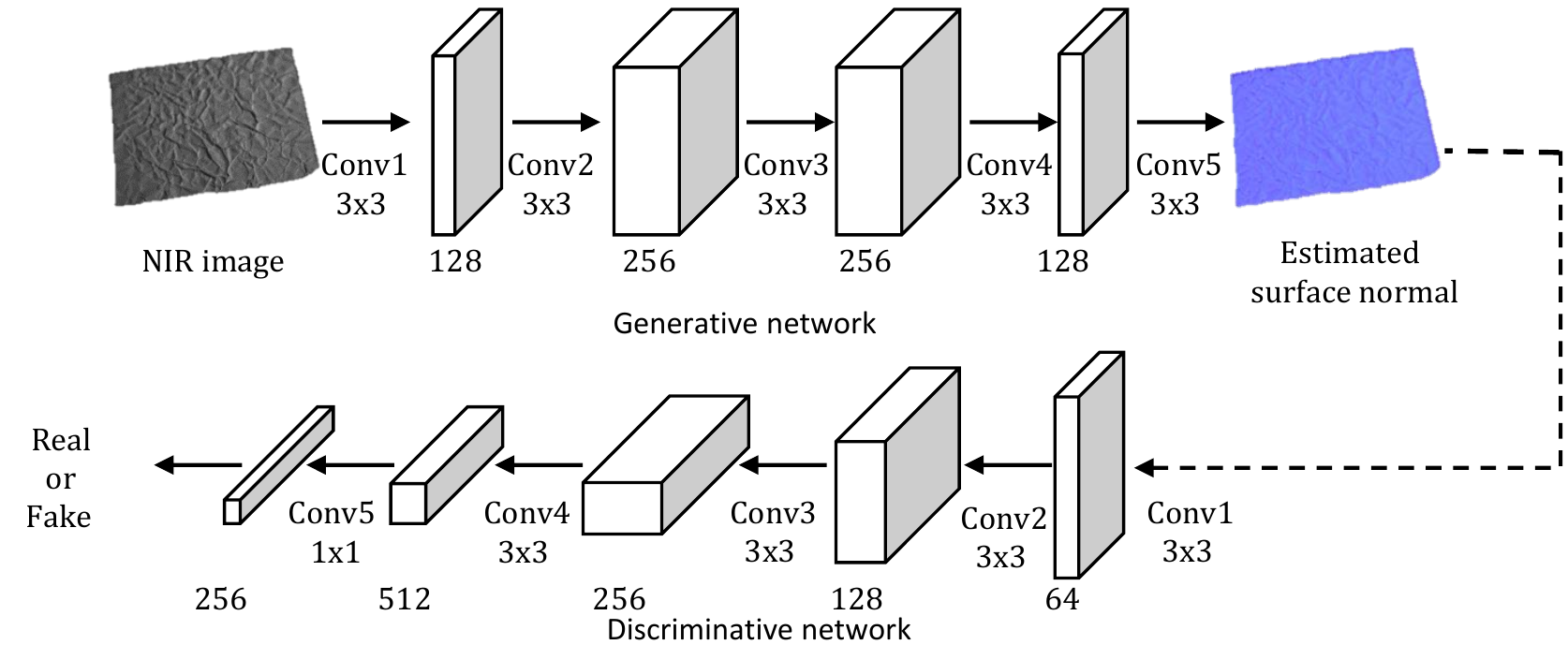}
\caption{Our network architecture. The proposed network produces surface normal map from a single NIR image. The generative model reconstructs surface normal map and the discriminative network predicts the probability whether the surface normal map comes from the training data or the generative model.}
\label{fig:Network}
\end{figure}

\subsection{Deep Shape from Shading}

Based on the generative adversarial network explained in \secref{sec:gan}, we modified the GAN model to be suitable for the shape-from-shading problem. Since shape-from-shading is the ill-posed problem, it is important to  incorporate proper constraints to uniquely determine the right solution. Therefore, we combine angular error and integrability loss, which are shown to be effective in many conventional SfS methods, into the objective function of the generative network.
Also, the existing GAN approaches typically take a random noise vector~\cite{goodfellow2014generative}, pre-encoded vector~\cite{radfor2015Unsupervisedrepresentation}, or an image~\cite{Mathieu2015deep-multi,denton2015deep} as the input of their generative networks, and each generative model produces the output which lies in the same domain as its input. 
In this work, we apply the generative model to produce a three-dimensional normal map from a NIR image where both data lies in the different domains. Compared to the conventional SfS methods, we do not need to calibrate the lighting directions. To the best of our knowledge, our work is the first application of the adversarial training to estimate fine-scale geometry from a single NIR image. 
Our network architecture is depicted in \figref{fig:Network}.

\subsubsection{Generative Networks} We use a fully convolutional network to construct the generative network. This type of a convolutional model was recently adopted in image restoration~\cite{Jiwon2015AccurateImage,dong2015image} and was verified its superior performance on the task. Typically, the convolutional operation squeezes the size of the input data. This relation, however, cannot be exploited to the full extent, so that it is not valid to generate images if the surround region is needed. To resolve this issue, we pad zeros before the convolution operation to keep the sizes of all feature maps including the output images. In the experiments, it turns out that this strategy works well in reconstructing the normal map.

\subsubsection{Discriminative Networks} Given the output of the generative network, a typical choice of the objectives function is the averaged $L_1$ or $L_2$ distance between ground-truth and generated output. However, such a choice has some limitations to be applied to our problem. $L_2$ distance produces blurry predictions because it assumes that the errors follow the Gaussian distribution. In $L_1$ distance, this effect could be diminished, but the estimated images would be the median of the set of equally likely intensities. 
We propose to add the discriminative network as a loss function with the distance metric. Recently, ~\cite{Mathieu2015deep-multi} proved that the combination of the distance, gradient and discriminative networks as a loss function provides the realistic and accurate output. Our discriminative model has a binary cross-entropy loss to make the high probability when the input is real images, and vice versa.

\subsection{Training}

\subsubsection{Generative Training} We will explain how we iteratively train the generative model $G$ and the discriminative model $D$. Let us consider a single NIR image $Z \in \{{Z_1,Z_2,...,Z_j}\}$ from a training dataset and the corresponding ground truth normal map $Y \in \{{Y_1,Y_2,...,Y_j}\}$. The training dataset covers various objects captured from diverse lighting directions, and we uniformly sampled the image from the dataset in terms of the balance of lighting directions. 

Basically, we followed the procedure of the paper~\cite{radfor2015Unsupervisedrepresentation}. Given $N$ paired image set, we first train $D$ to classify the real image pair $(Z, Y)$ into the class $1$ and the generated pair $(Z, G(Z))$ into the class $0$. In this step, we fixed the parameters $(\theta_G)$ of the generative network $G$ to solely update the parameters $(\theta_D)$ of $D$. The objective function of the discriminative model is denoted as: 
\begin{equation}
\mathcal{L}_{D} (Z,Y) = \sum_{i=1}^{N}~\mathcal{D}_{bce}(Y_i,1) +\\ \mathcal{D}_{bce}(G(Z_i),0),
%	\min_{\Theta_G}\max_{\Theta_D}\\
%	\displaystyle \E_{F^i\sim D_{desire}}[logD( I^i )] + \\
%	\displaystyle \E_{Z^j\sim D_{input}}[log(1 - D(F^i) ))]
%\frac{1}{N}\sum_{i=1}^{N}\mathcal{L}_D(I^i)+\frac{1}{N}\sum_{j=1}^{N}\mathcal{L}_D( G(Z^j) ),\\
\end{equation}
where $\mathcal{D}_{bce}$ is the binary cross-entropy, defined as 
\begin{equation}
\mathcal{D}_{bce}(Y_i,C) = - C_i log(Y_i) + (1-C_i)log(1-Y_i),
%\mathcal{L}_{D} (Z,Y) = \sum_{i=1}^{N}~\mathcal{D}_{bce}(Z_1^i,Y_1^i,1) +\\ %\mathcal{D}_{bce}(Z_1^i,G(Z_1^i),0),
%	\min_{\Theta_G}\max_{\Theta_D}\\
%	\displaystyle \E_{F^i\sim D_{desire}}[logD( I^i )] + \\
%	\displaystyle \E_{Z^j\sim D_{input}}[log(1 - D(F^i) ))]
%\frac{1}{N}\sum_{i=1}^{N}\mathcal{L}_D(I^i)+\frac{1}{N}\sum_{j=1}^{N}\mathcal{L}_D( G(Z^j) ),\\
\end{equation}
where $C_i$ is the binary class label. We minimize the objective function toward the state to be assigned high probability scores into real images $Y_i$ and low probability scores into generated images $G(Z_i)$.

After that, we keep the parameters of $D$ fixed and train the generative model $G$.
Many previous deep learning based image restoration and generation methods~\cite{Jiwon2015AccurateImage,John2015deepstereo} used the mean square error(MSE) loss function to minimize between the ground-truth images and output images. However, as studied in the conventional SfS works, estimating accurate surface normal maps requires to minimize angular errors and the output normals satisfy the integrability constraint. Therefore, we modified the objective function of the GAN model to incorporate those photometric objective functions. By taking the objective functions, we can effectively remove angular error and estimate physically meaningful surface normals.

Specifically, to evaluate surface normal properly, we defined the objective function of our generative network as:
\begin{equation}
\mathcal{L}_{G} (Z,Y) = \sum_{i=1}^{N}~\mathcal{D}_{bce}(G(Z_i),1) + \lambda_{l_p}L_p + \lambda_{ang}L_{ang} + \lambda_{curl}L_{curl}. %||\bigtriangledown \times G(Z) \rangle||^n
\end{equation}

Following the conventional $L_1$ or $L_2$ loss, the estimated normal map difference $\mathcal{L}_{p}$ is denoted as: 
\begin{equation}
\mathcal{L}_{p}(Y,G(Z)) =||Y-G(Z)||_p^p 
%\mathcal{L}_{ang}(Y,G(Z)) =1 - \langle Y, G(Z) \rangle = 1 - \frac{Y^TG(Z)}{\parallel{Y}\parallel \parallel{G(Z)}\parallel}  
\end{equation}
where $p=1$ or $p=2$

To estimate the accuracy of photometric stereo, the angular error is often used in the conventional photometric approaches because it describes more physically meaningful error than direct normal map difference. To minimize the angular error, we normalize both the estimated normals ($G(Z)$) and the ground-truth normals ($Y$), then simply apply the dot product between them as:
\begin{equation}
\mathcal{L}_{ang}(Y,G(Z)) =1 - \langle Y, G(Z) \rangle = 1 - \frac{Y^TG(Z)}{||Y|| ||G(Z)||}  
%\mathcal{L}_{ang}(Y,G(Z)) =1 - \langle Y, G(Z) \rangle = 1 - \frac{Y^TG(Z)}{\parallel{Y}\parallel \parallel{G(Z)}\parallel}  
\end{equation}

The angular error provides physically meaningful measures, however it averaged entire surface normals. In order to encourage the generative network to estimate photometrically correct surface normals, we also add the integrability constraint in local neighbors into the objective function, which is denoted as:
\begin{equation}
\mathcal{L}_{curl} = ||\bigtriangledown \times G(Z) \rangle||.
\end{equation}
The integrability constraint enforces that the integral of normal vectors in a local closed loop must sum up to zero, meaning that angles are returned to the same height. The integrability constraint prevents a drastic change and guarantees estimated normals lie on the same surface in a local region.

\section{Experiment}
\subsection{Dataset}
To apply deep learning framework to our purpose, it is required to correct the numerous and good quality dataset. However, many existing dataset are not enough to train the network and also belonging targets are often inadequate for our tasks. Recently Choe \etal~\cite{Choe16cvpr} opened a new NIR benchmark dataset, including 101 real-world objects such as fabrics, leaves and paper taken at 9 viewings and 12 lighting directions.

We used a pair of NIR as input and surface normal maps as target for ground truth. For fine-scale refinement, we augmented NIR images into 12 patches ($64\times64$) within a single ground truth. For training, we used images from 91 objects and the remaining objects are for validation and test dataset. Note that we uniformly sampled validation and test samples according to the object category.  

(Before: For training, we use surface normal maps from 12 image photometric stereo that the surface normal maps from \cite{Choe16cvpr} is ground-truth therefore 12 NIR images are shared one surface normal map ground-truth. We train our network with $64\times64$ size of NIR patches from 91 objects and the remaining of objects are used for validation. The validation sets are evenly selected considering the different categories of the entire dataset.)

When we trained the network, we normalized NIR images and normal maps to $[-1, +1]$.

\begin{figure}
\centering
\begin{tabular}{c@{\hspace{1mm}}c@{\hspace{1mm}}}
\includegraphics[width=0.99\linewidth]{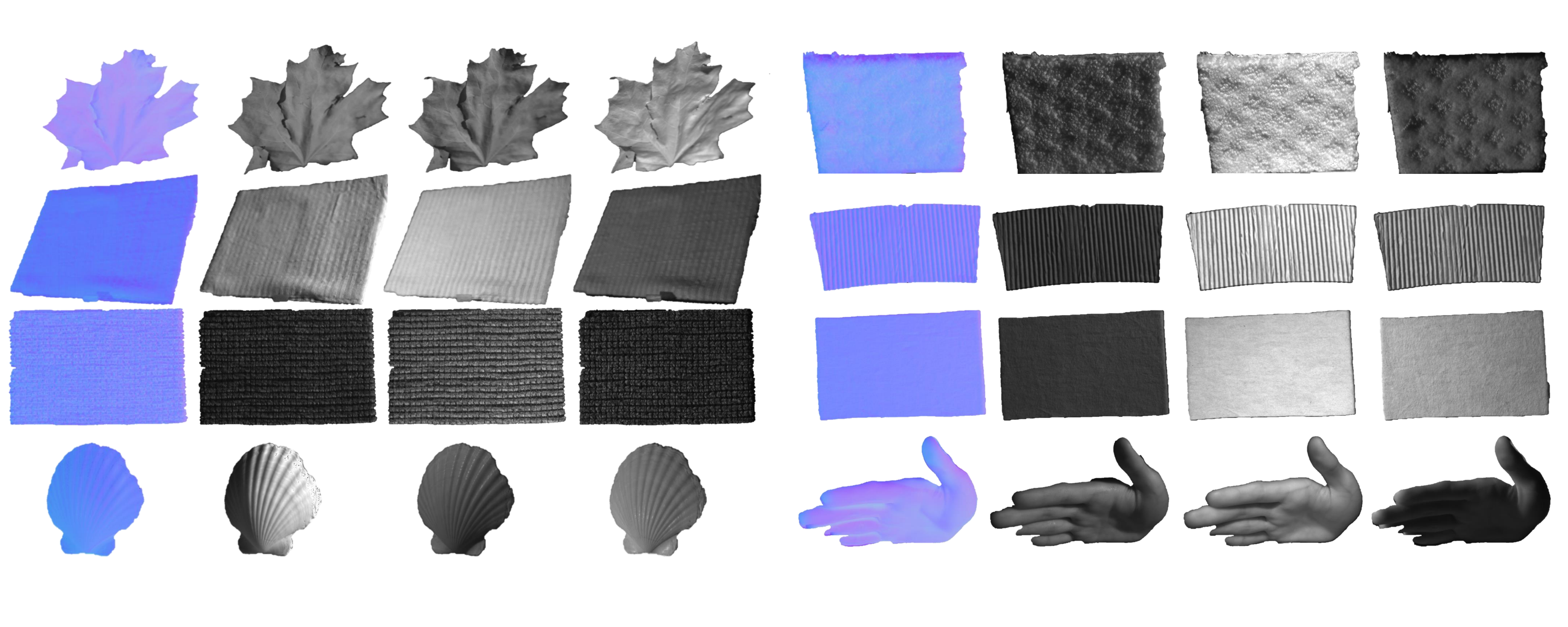}
\end{tabular}
\caption{ Dataset~\cite{Choe16cvpr} has various real-world object taken by 12 different lighting directions and 9 objects of view points. The leftmost is a normal map as the ground-truth and others are NIR images from different lighting directions. The Variety of lighting directions makes the same object appear vastly different.} 

\label{fig:trainingdata}
\end{figure}

\subsection{Training Parameters}

We provide parameters used to train our proposed network. The configuration of the network is depicted in~\tabref{tab:arch}. Training used batches of size 32. For initializing weights, we assigned a Gaussian distribution with a zero mean and $0.02$ of a standard deviation. We trained all experiments over $46000$ iterations to use Adam optimizer~\cite{kingma2014adam} with momentum$~\beta_1 =0.5.$ The learning rate was starting to $0.0002$, decreased by a factor of $0.95$ every $5000$ iterations. For balancing the scale of normalization, we set a hyperbolic tangent at the end of the generative network. Lastly, we adjusted  $5\times5$ of local neighboring with 3 pixels overlap to compute the integrability. 
In optimization procedure, we used a combined loss function including intensity($L_p$), angular($L_{ang}$), and integrability constraint($L_{curl}$). Note that we did not tune the weighted parameters of each loss functions so that we used the same weighting as $~\lambda_p = \lambda_{ang} = \lambda_{curl}$ =1.

\begin{table}[t]
	\centering
	\begin{tabular}{|c|c|c|c|c|c|c|}\hline
		Layer&\bigcell{c}{Number\\of filters}&\bigcell{c}{Filter size\\(w$\times$h$\times$ch)}&Stride&Pad&\bigcell{c}{Batch\\norm.}&\bigcell{c}{Activation\\function}\\\hline\hline
		Conv. 1&64&3$\times$3$\times$3&2&0&$\times$&L-ReLU\\
		Conv. 2&128&3$\times$3$\times$64&2&0&$\ocircle$&L-ReLU\\
		Conv. 3&256&3$\times$3$\times$128&2&0&$\ocircle$&L-ReLU\\
		Conv. 4&512&3$\times$3$\times$256&2&0&$\ocircle$&L-ReLU\\
		Conv. 5&256&1$\times$1$\times$512&1&0&$\times$&sigmoid\\\hline
	\end{tabular}\\
	(a) Details of the Discriminative network.\\$\;$\\
	
	\begin{tabular}{|c|c|c|c|c|c|c|}\hline
		Layer&\bigcell{c}{Number\\of filters}&\bigcell{c}{Filter size\\(w$\times$h$\times$ch)}&Stride&Pad&\bigcell{c}{Batch\\norm.}&\bigcell{c}{Activation\\function}\\\hline\hline
		Conv. 1&128&3$\times$3$\times$32&1&1&$\ocircle$&ReLU\\
		Conv. 2&256&3$\times$3$\times$128&1&1&$\ocircle$&ReLU\\
		Conv. 3&256&3$\times$3$\times$256&1&1&$\ocircle$&ReLU\\
		Conv. 4&128&3$\times$3$\times$256&1&1&$\ocircle$&ReLU\\
		Conv. 5&3&3$\times$3$\times$128&1&1&$\times$&tanh\\\hline
	\end{tabular}\\
	(b) Details of the Generative network.\\$\;$\\
	\caption{Network Configuration.}
	\label{tab:arch}
\end{table}

\subsection{Experimental Result}
We use Tensorflow~\cite{Tensorflow} to implement and train the proposed network. The proposed network is fully convolutional network, we apply the entire NIR image at evaluation. Computation time to estimate a surface normal is about 2 seconds with Titan X, meanwhile the conventional shaped from shading method takes 10 minutes with Matlab implementation.\\

\subsubsection{Quantitative Analysis} 

For the quantitative evaluation, firstly, we validate the each terms of our cost functions. In this experiment, we tested our method using 3rd NIR direction among 12 lighting directions. To evaluate the performance of our method, we use three metrics; angular error, good pixel ratio and intensity error. In ~\tabref{tab:error}, all the quantitative errors are shown. Compared to case of using only intensity loss, when the angular cost function added, the performance is improved. This validates that our angular loss measures the physically meaningful error. The integrability term insures the continuity of the local normals. Although the integrability is satisfied for most of smooth surfaces, it does not guarantee performance improvement in some non-smooth surfaces. In our experiments, $L_2 + L_{ang}$ loss function shows the best performance for all views case, and $L_1 + L_{ang}$ achieves the lowest error for center view case. We compare our results with the conventional SfS method and we verified that our framework competitively performs. We also compare our method with the deep CNN-based surface normal estimation method~\cite{eigen2015predicting}. Although this method estimates the surface normal, it is designed for reconstructing the scene-level low-frequency geometries and is not suitable for our purpose.    
We also measure errors for the center view (5th viewing direction). Since extreme viewing directions are saturated or under-exposed in some cases, measuring the error of the center view results in lower errors. We found that estimated normal maps are distorted in extreme view points~(error in low-frequency geometry). To evaluate the fine-scale (High-frequency) geometry, we define a detail map ($M$) based on the measure in~\cite{nehab2005efficiently}. This measure is computed as:  $M = f(Y) + G(Z) - f(G(Z))$, where function $f$ is smoothing function.

\begin{table*}[t]
	\centering
	\begin{tabular}{|l|l|c|c|c|c|c|c|c|}\hline
		\multicolumn{2}{|c|}{}&\multicolumn{2}{c|}{Angular error($^\circ$)}&\multicolumn{3}{c|}{Good pixels(\%)}&\multicolumn{2}{c|}{Intensity error} \\ 
		\multicolumn{2}{|c|}{}&\multicolumn{2}{c|}{(Lower Better)}&\multicolumn{3}{c|}{(Higher Better)}&\multicolumn{2}{c|}{(abs error)} \\ \cline{1-9}
		\multicolumn{1}{|c|}{View points}&\multicolumn{1}{|c|}{Methods}&Mean&Median&10$^\circ$&15$^\circ$&20$^\circ$&Mean&Median \\ \hline\hline
		\multirow{7}{*}{All views}&$L_1$&\multicolumn{1}{c}{16.82}&\multicolumn{1}{c}{16.18}&\multicolumn{1}{c}{17.10}&\multicolumn{1}{c}{38.23}&\multicolumn{1}{c}{72.60}&\multicolumn{1}{c}{0.14}&\multicolumn{1}{c|}{0.09} \\
		&$L_2$&\multicolumn{1}{c}{16.72}&\multicolumn{1}{c}{16.68}&\multicolumn{1}{c}{17.49}&\multicolumn{1}{c}{36.12}&\multicolumn{1}{c}{69.80}&\multicolumn{1}{c}{0.14}&\multicolumn{1}{c|}{0.09} \\
		&$L_1$+$L_{ang}$&\multicolumn{1}{c}{15.88}&\multicolumn{1}{c}{15.80}&\multicolumn{1}{c}{19.39}&\multicolumn{1}{c}{37.13}&\multicolumn{1}{c}{73.08}&\multicolumn{1}{c}{0.13}&\multicolumn{1}{c|}{0.09} \\
		&$L_2$+$L_{ang}$&\multicolumn{1}{c}{\textbf{15.56}}&\multicolumn{1}{c}{\textbf{15.31}}&\multicolumn{1}{c}{20.26}&\multicolumn{1}{c}{\textbf{49.77}}&\multicolumn{1}{c}{\textbf{74.39}}&\multicolumn{1}{c}{0.13}&\multicolumn{1}{c|}{0.08} \\		
		&$L_1$+$L_{ang}$+$L_{curl}$&\multicolumn{1}{c}{16.46}&\multicolumn{1}{c}{16.35}&\multicolumn{1}{c}{\textbf{23.32}}&\multicolumn{1}{c}{41.46}&\multicolumn{1}{c}{65.43}&\multicolumn{1}{c}{0.14}&\multicolumn{1}{c|}{0.09} \\		
		&$L_2$+$L_{ang}$+$L_{curl}$&\multicolumn{1}{c}{15.71}&\multicolumn{1}{c}{15.49}&\multicolumn{1}{c}{18.56}&\multicolumn{1}{c}{41.97}&\multicolumn{1}{c}{73.46}&\multicolumn{1}{c}{0.14}&\multicolumn{1}{c|}{0.09} \\	
		&Detail    
		  map ($L_2$+$L_{ang}$)&\multicolumn{1}{c}{\textbf{3.61}}&\multicolumn{1}{c}{\textbf{2.99}}&\multicolumn{1}{c}{\textbf{95.98}}&\multicolumn{1}{c}{\textbf{99.10}}&\multicolumn{1}{c}{\textbf{99.61}}&\multicolumn{1}{c}{\textbf{0.06}}&\multicolumn{1}{c|}{\textbf{0.02}} \\	
		\hline
		\multirow{7}{*}{Center view}
		&$L_1$&\multicolumn{1}{c}{10.01}&\multicolumn{1}{c}{9.19}&\multicolumn{1}{c}{58.17}&\multicolumn{1}{c}{82.82}&\multicolumn{1}{c}{93.47}&\multicolumn{1}{c}{0.08}&\multicolumn{1}{c|}{0.05} \\
		&$L_2$&\multicolumn{1}{c}{8.76}&\multicolumn{1}{c}{8.37}&\multicolumn{1}{c}{67.14}&\multicolumn{1}{c}{90.97}&\multicolumn{1}{c}{97.44}&\multicolumn{1}{c}{0.07}&\multicolumn{1}{c|}{0.05} \\
		&$L_1$+$L_{ang}$&\multicolumn{1}{c}{7.35}&\multicolumn{1}{c}{6.74}&\multicolumn{1}{c}{77.07}&\multicolumn{1}{c}{93.90}&\multicolumn{1}{c}{98.58}&\multicolumn{1}{c}{0.06}&\multicolumn{1}{c|}{0.04} \\
		&$L_2$+$L_{ang}$&\multicolumn{1}{c}{7.70}&\multicolumn{1}{c}{6.82}&\multicolumn{1}{c}{73.36}&\multicolumn{1}{c}{91.90}&\multicolumn{1}{c}{98.35}&\multicolumn{1}{c}{0.07}&\multicolumn{1}{c|}{0.04} \\		
		&$L_1$+$L_{ang}$+$L_{curl}$&\multicolumn{1}{c}{9.94}&\multicolumn{1}{c}{9.42}&\multicolumn{1}{c}{54.90}&\multicolumn{1}{c}{85.24}&\multicolumn{1}{c}{96.07}&\multicolumn{1}{c}{0.09}&\multicolumn{1}{c|}{0.05} \\		
		&$L_2$+$L_{ang}$+$L_{curl}$&\multicolumn{1}{c}{\textbf{4.75}}&\multicolumn{1}{c}{\textbf{4.05}}&\multicolumn{1}{c}{\textbf{93.47}}&\multicolumn{1}{c}{\textbf{98.87}}&\multicolumn{1}{c}{\textbf{99.53}}&\multicolumn{1}{c}{0.06}&\multicolumn{1}{c|}{0.03} \\		
        &Detail    
        map ($L_1$+$L_{ang}$)&\multicolumn{1}{c}{\textbf{4.14}}&\multicolumn{1}{c}{\textbf{3.67}}&\multicolumn{1}{c}{\textbf{95.95}}&\multicolumn{1}{c}{\textbf{99.53}}&\multicolumn{1}{c}{\textbf{99.84}}&\multicolumn{1}{c}{0.05}&\multicolumn{1}{c|}{0.02} \\	
        \hline\hline
		\multirow{2}{*}{Center view}&\multicolumn{1}{l|}{SfS (Detail map)}&\multicolumn{1}{c}{5.09}&\multicolumn{1}{c}{4.14}&\multicolumn{1}{c}{88.25}&\multicolumn{1}{c}{97.19}&\multicolumn{1}{c}{99.27}&\multicolumn{1}{c}{0.06}& \multicolumn{1}{c|}{0.03}\\
		\multicolumn{1}{|c|}{}&\multicolumn{1}{l|}{Eigen~\etal~\cite{eigen2015predicting}}&\multicolumn{1}{c}{77.87}&\multicolumn{1}{c}{80.78}&\multicolumn{1}{c}{0.48}&\multicolumn{1}{c}{0.96}&\multicolumn{1}{c}{1.52}&\multicolumn{1}{c}{0.61}& \multicolumn{1}{c|}{0.75}\\
	    \hline
			
	\end{tabular}
	\caption{Quantitative evaluation. We validate the each terms of our cost functions with various error measures.}
	\label{tab:error}
\end{table*}

\subsubsection{Qualitative Analysis}
\begin{figure}
	\centering
	\begin{tabular}{c@{\hspace{1mm}}c@{\hspace{1mm}}}
		\includegraphics[width=0.99\linewidth]{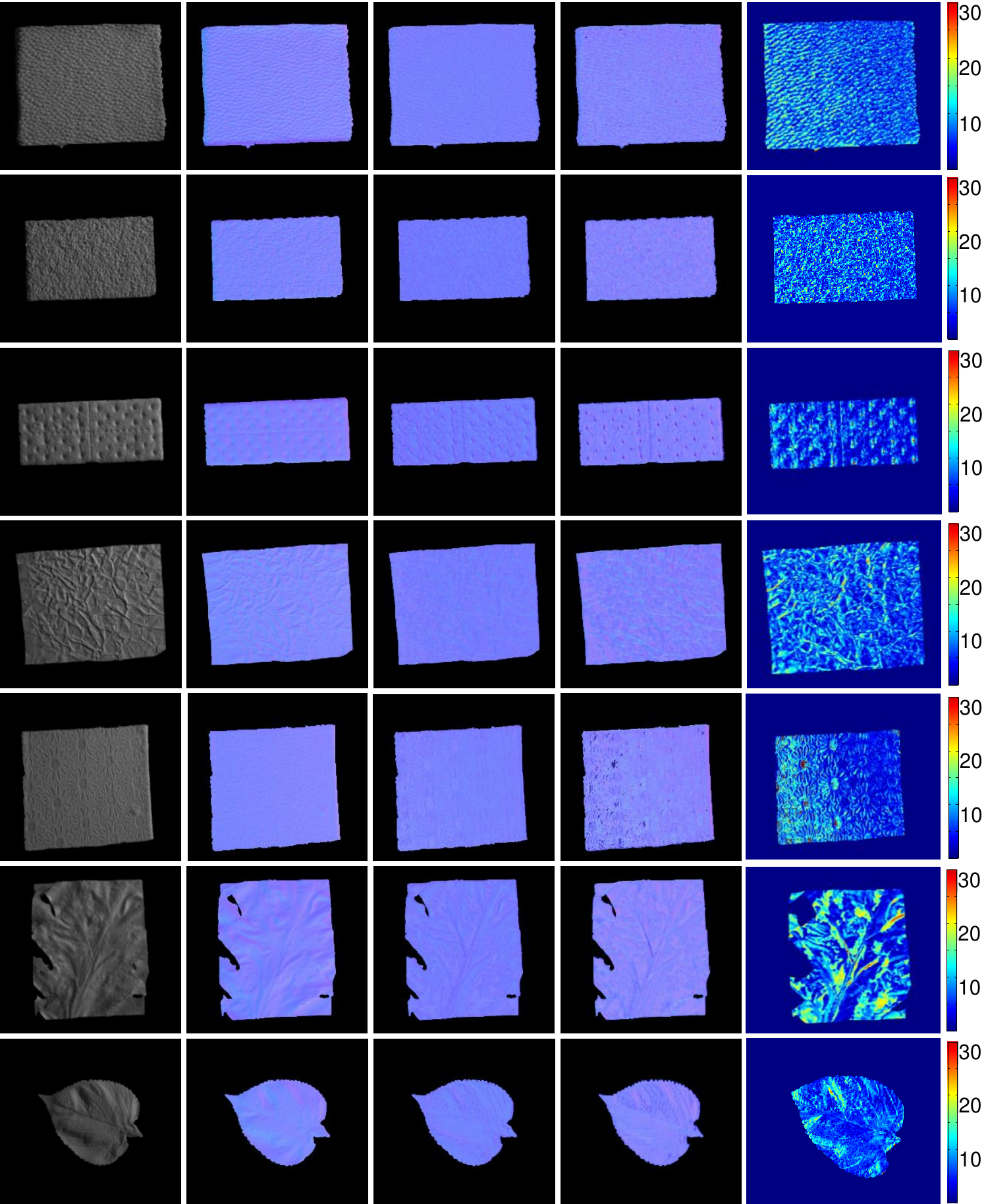}
	\end{tabular}
	\caption{Qualitative results of surface normal estimation using the proposed network. From left to right: Input NIR images, ground-truth, normal from $L_2$, $L_2+L_{ang}$ and error map between ground-truth and $L_2 + L_{ang}$. }
	\label{fig:normal_result}
\end{figure}

\begin{figure}
\centering
\begin{tabular}{c@{\hspace{1mm}}c@{\hspace{1mm}}}
\includegraphics[width=0.99\linewidth]{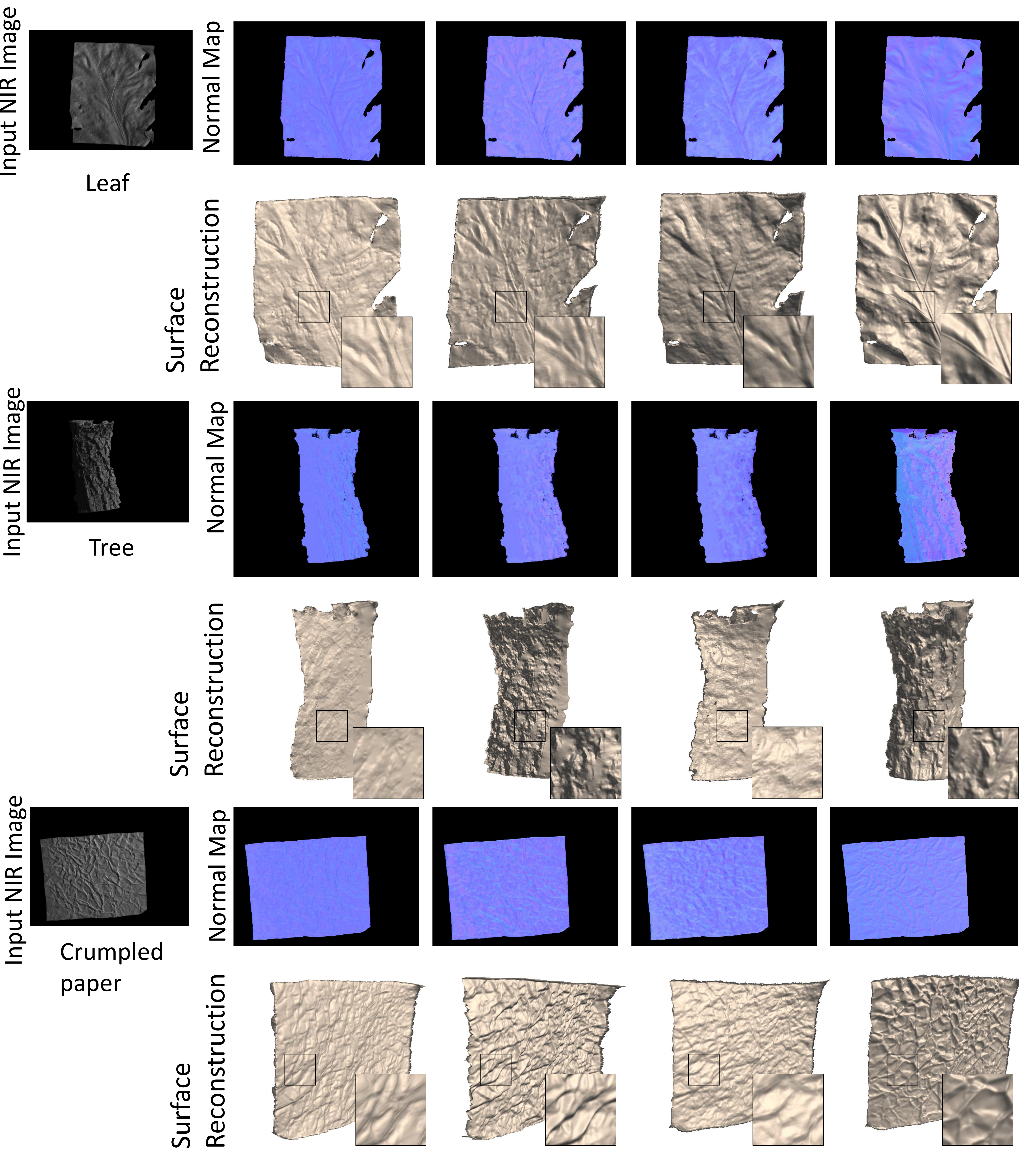}
\end{tabular}
\caption{ Surface reconstruction results. From left to right: input, $L_2$, $L_2 + L_{ang}$, $L_2 + L_{ang} + L_{curl}$ and ground-truth. We compute depth from the surface normal and reconstruct mesh. All three cases are visualized.}
\label{fig:3dmodel}
\end{figure}

~\figref{fig:normal_result} and ~\figref{fig:3dmodel} show the qualitative results of our network. Our network is able to estimate fine-scale textures of objects. Comparing between $L_2$ and $L_2 + L_{ang}$, we figure out that the angular loss provides more fine-scale textures than intensity loss. By adding the integrability constraint, the result produces a smoother surface. This demonstrates, therefore, that our network is trained to follow physical properties relevant to SfS.

\subsection{Shape Estimation at Arbitrary Lighting Direction}
\begin{figure}[t]
	\centering
	\begin{tabular}{c@{\hspace{1mm}}c@{\hspace{1mm}}}
		\includegraphics[width=0.99\linewidth]{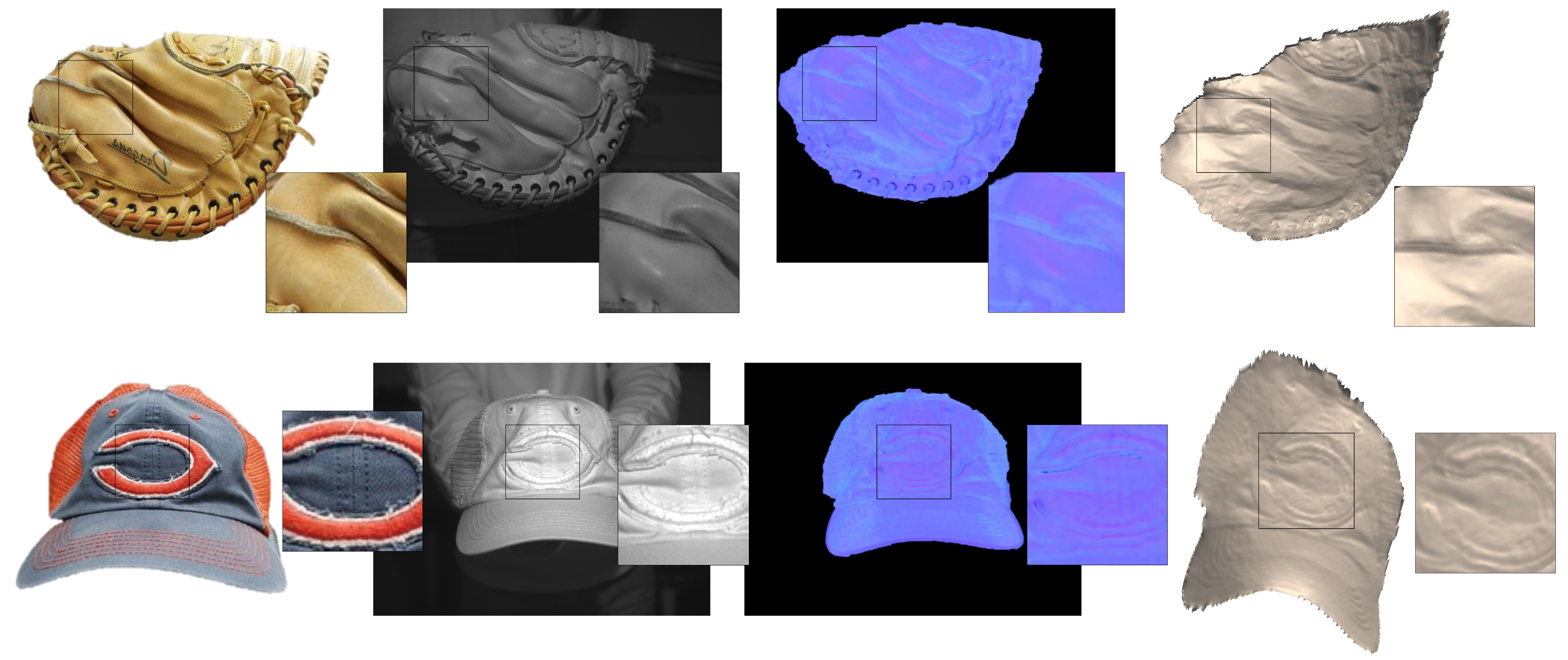}
	\end{tabular}
	\caption{Surface normal reconstruction results from an arbitrary lighting direction. From left to right, the columns show the RGB images, NIR images, estimated surface normals, and reconstructed 3D models.}
	\label{fig:Arbitrary}
\end{figure}

We evaluate our network for the surface estimation with an arbitrary lighting direction. Without prior knowledge of the lighting directions, SfS becomes a more challenging problem. As shown in~\figref{fig:Arbitrary}, we captured several real-world objects. The glove has a complex surface geometry. Note that the bumpy surface and the stitches at the bottom are reconstructed. 
The cap has a 'C' letter on it and the geometry of this is reconstructed in mesh result.

\section{Future Work}
Our approach reconstructs high-frequency detail map. However, in some cases in our experiments, we observed that some distortions of low-frequency geometry occur. To deal with this issue, in future, our network can be extended to combine the conventional rough geometry estimation methods which reconstruct the low-frequency geometries.

\section{Conclusion}
In this paper, we have presented a generative adversarial network for estimating surface normal maps from a singel NIR image. As far as we aware, this is the first work to estimate fine-scale surface geometry from a NIR images using a deep CNN framework. The proposed network shows competitive performance without any lighting information. We demonstrated that our photometically-inspired object function improves the quality of surface normal estimation. We also applied our network to arbitrary NIR images which are captured under different configuration with the training dataset and have shown the promising results.

\bibliographystyle{splncs}
\bibliography{egbib}
\end{document}